\documentclass[sigconf]{acmart}
\usepackage{hyperref}
\usepackage{graphics}
\usepackage{booktabs}
\AtBeginDocument{%
  \providecommand\BibTeX{{%
    \normalfont B\kern-0.5em{\scshape i\kern-0.25em b}\kern-0.8em\TeX}}}

\setcopyright{acmcopyright}
\copyrightyear{2021}
\acmYear{2021}
\acmDOI{10.1145/1122445.1122456}

\begin{document}

\title{The Role of Facial Expressions and Emotion in ASL}

\author{Lee Kezar}
\author{Pei Zhou}

\affiliation{%
  \institution{University of Southern California}
  \country{United States of America}
}

\renewcommand{\shortauthors}{Pei Zhou and Lee Kezar}



\begin{abstract}
There is little prior work on quantifying the relationships between facial expressions and emotionality in American Sign Language. In this final report, we provide two methods for studying these relationships through probability and prediction. Using a large corpus of natural signing manually annotated with facial features paired with lexical emotion datasets, we find that there exist many relationships between emotionality and the face, and that a simple classifier can predict what someone is saying in terms of broad emotional categories only by looking at the face.
\end{abstract}


\maketitle

\section{Introduction}

It is well established that the face plays an important role in human communication. Actions such as wrinkling the nose, squinting the eyes, smiling, and raising the eyebrows have cultural and linguistic ties in everyday communication, either emphasizing or adding information to the verbal register. This is especially true for American Sign Language, where meaningful gestures include not only the hands, but also the arms, torso, and head.
Indeed, facial features (i.e. non-manual markers) can have a grammatical or semantic role, functioning similarly to vocal features (e.g. pitch, volume) in speech \citep{reilly:affective_prosody}.

However, there is no prior work on studying the exact meanings of these facial expressions. It is usually assumed that most facial expressions are common-sense, such as smiling to convey happiness and squinting eyes to convey confusion or anger. This is misleading, as ASL contains many arbitrary, idiosyncratic facial expressions that are not common in other languages, such as puffing one's cheeks to show large and bloated objects, or looking to the left or right to refer to a third person.

These examples, which are semantic and linguistic in nature, demonstrate that a more nuanced multimodal analysis of facial expressions is necessary to fully understand ASL. In this project, we seek to provide evidence of well-known relationships between the face, ASL grammar, and semantics. In particular, we focus on signs and facial expressions that work together to convey \textit{emotion}. We also seek to expand our knowledge of the role of facial expressions and ASL by uncovering new relationships. Finally, we contribute multimodal data for a corpus of ASL that can be used for future investigations into the role of facial features in ASL.

\section{Method}
We evaluate the role of facial expressions in ASL by finding relationships among a rich set of multimodal features. These features are comprised of lexical and facial emotion (e.g. \textit{surprised}), facial action units (e.g. \textit{inner brow raisier}, \citep{ekman:facs_manual}), and linguistic markers (e.g. \textit{rhetorical question}).

\paragraph{Note on Terminology} Within each category are multiple \textit{types} and \textit{values}. For example, within the category of facial action units, there are feature \textit{types} pertaining to the eyebrows, eyes, nose, mouth, etc. Within each of these, there are a few discrete \textit{values} that the feature can take, such as \textit{mouth open} and \textit{eyebrows raised}. This distinction is useful for finding general patterns between feature types, in addition to finer-grained patterns between values.

\subsection{Linguistic Modality}
The linguistic modality contains ASL gloss (a sequence of identified signs, analogous to transcribed speech), syntactic descriptors (such as negation, rhetorical questions, and role shifting), and English translation. We complement this data by including emotional features based on word choice (more in section \ref{emotion_feats}).

\subsection{Visual Modality}
The visual modality contains videos of people signing and manually-tagged facial features (such as eyebrows raised or mouth open).

We complement this data by including a facial expression recognition model\footnote{\url{https://pypi.org/project/fer/}} trained on FER 2013 dataset~\cite{goodfellow2013challenges} and uses an MTCNN~\cite{zhang2016joint}. The input of the model is an image and the output is a distribution score over 7 emotion classes: angry, disgust, fear, happy, sad, surprise, and neutral.

\subsection{Multi-modal Analysis}
\paragraph{Co-occurrence}
In the first phase of this project, we explore the inter-modal relationships between a facial features and lexical features. We compute the conditional probability of a frame containing feature $a$ given the co-occurrence of feature $b$ using the simple formula $ P(a | b) = \frac{\#(a, b)}{\#(a)} $.
\paragraph{Multi-label Classification of Emotions}
In the second phase, we explore the relationships between joint facial features and lexical features. We train a Random Forest classifier to predict lexical emotion based on a one-hot encoding of facial action units for each frame. In doing so, we empirically determine to what extent the face \textit{as a whole} communicates affect.

\section{Data}

\subsection{SignStream}
The National Center for Sign Language and Gesture Resources (NCSLGR)'s SignStream corpus \citep{neidle:signstream} contains 67 videos of native ASL users signing naturally. It also contains detailed annotations about the signer's facial expressions and the signs they are producing.

We preprocess this data by extracting frames for all videos and aligning them with their annotations. Features that span multiple frames are copied such that the features for any frame can be easily retrieved.

Then, using SignStream's English translation data, we find all emotion features present in the text using the process in Section \ref{emotion_feats}. The resulting data contains 180,669 frames with associated emotion labels.

\begin{figure}[h!]
    \centering
    \includegraphics[width=0.45\textwidth]{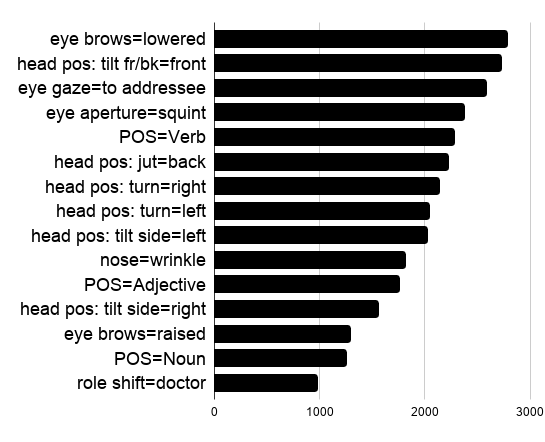}
    \caption{The frame counts for SignStream features}
    \label{fig:face_counts}
\end{figure}

\begin{figure}[h!]
    \centering
    \includegraphics[width=0.45\textwidth]{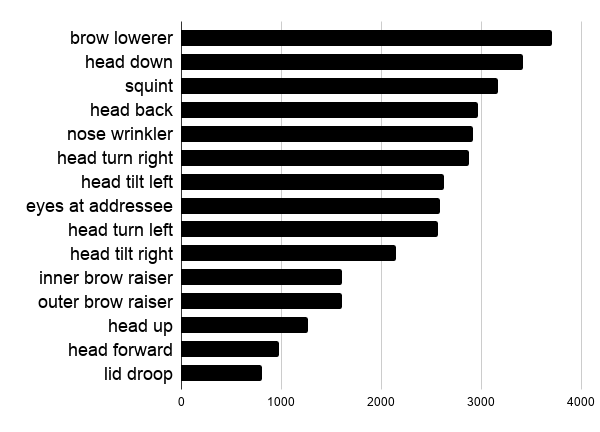}
    \caption{The frame counts for Facial Action Units.}
    \label{fig:fau_counts}
\end{figure}

Finally, we map SignStream facial features to the Facial Action Coding System (e.g. "eye brows lowered" becomes "AU 4"). The counts are shown in Figures \ref{fig:face_counts} and \ref{fig:fau_counts}. The distributions differ because some facial action units have multiple SignStream features, such as eyebrows "slightly lowered" and "further lowered".

\subsection{Lexical Emotion Features} \label{emotion_feats}
The Linguistic Inquiry and Word Count (LIWC \citep{tausczik:liwc}) and Empath \citep{fast:empath} provide word-to-feature lexicons that span meaning, syntax, emotionality, and cognition. For this project, we use emotional features only, although future work might explore the role of semantics more generally.

\begin{figure}[h!]
    \centering
    \includegraphics[width=0.45\textwidth]{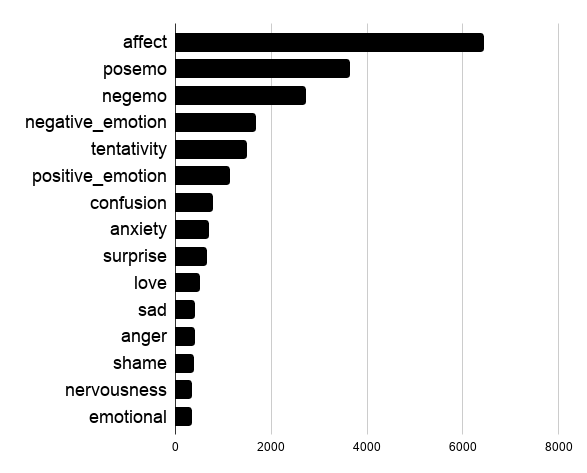}
    \caption{The frame counts for lexical emotion features}
    \label{fig:emotion_counts}
\end{figure}

The emotional features are obtained by checking all tokens in the English translation for potential emotional words (as defined by the lexica). The counts are shown in Figure \ref{fig:emotion_counts}. We eliminate features that contain less than 10 instance frames, resulting in 18 lexical features.

\subsection{Synthesis}
For each frame in the SignStream corpus, we automatically extract the current sign, its lexical features, and all features related to the face. In total, this yields 6,390,883 features across 180,669 frames for an average of 35 features per frame. This highly expressive, multimodal dataset is publicly available.\footnote{https://bit.ly/33ys9W4}

\section{Results}
\subsection{Co-occurrence Analysis}
\begin{figure}
    \centering
    \includegraphics[width=0.45\columnwidth]{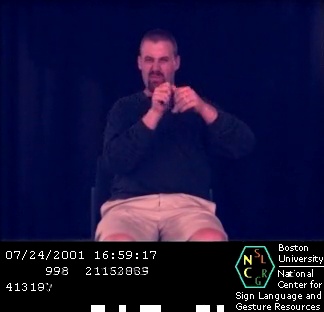}
    \includegraphics[width=0.45\columnwidth]{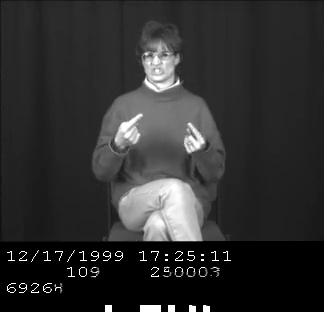}
    \caption{Examples of negative emotion signs ("cut", left, and "hurt", right) relying on lowered eyebrows and wrinkled nose.}
    \label{fig:negemo_example}
\end{figure}

For each of the 9697 frames in our created dataset, we compute the conditional probability of a feature $a$ given $b$ using their co-occurrence statistic. We present preliminary findings on potential relationships shown from our conditional probabilities. Table~\ref{fig:cond_prob_negemo} shows the top 10 features extracted the SignStream data that has the highest conditional probability with the emotion label ``negative emotion'' (we removed too infrequent ones). We can see that nose wrinkle and lowered eye brows are potentially correlated with the negative affect more (examples in figure \ref{fig:negemo_example}).

\begin{figure}
    \centering
    \includegraphics[width=0.45\columnwidth]{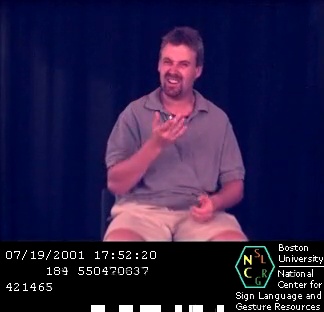}
    \includegraphics[width=0.45\columnwidth]{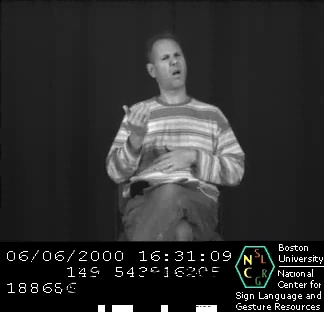}
    \caption{Examples of confusion emotion signs ("why?" in both) relying on lowered eyebrows and tensed cheeks.}
    \label{fig:confusion_example}
\end{figure}

Table~\ref{fig:cond_prob_confusion} shows the top 10 features extracted the SignStream data that has the highest conditional probability with the emotion label ``confusion'' . We can find intuitive features such as asking a question and tensed cheeks (examples in figure \ref{fig:confusion_example}).

As we can see in the examples, the primary difference between these two emotions is the mouth: confusion tends to rely on an open mouth while negative emotion tends to rely on a closed mouth. This is affirmed with the data

\begin{table}[h]
\resizebox{\columnwidth}{!}{
\begin{tabular}{c|c}
\toprule 
\multicolumn{1}{c}{
\textbf{Feature}} & \textbf{Conditional Probability} \\
\midrule 
nose=wrinkle                         & 0.694                            \\
head mvmt: jut                       & 0.644                            \\
Non-dominant POS=Classifier          & 0.638                            \\
eye gaze=to addressee                & 0.605                            \\
topic/focus=focus/top1               & 0.500                          \\
head mvmt: side to side=slow         & 0.494                            \\
head pos: tilt fr/bk=front           & 0.469                            \\
cheeks=puffed                        & 0.434                            \\
eye brows=lowered                    & 0.434                            \\
head pos: jut=back                   & 0.409                           \\
\bottomrule
\end{tabular}
}
\caption{Top ten associated features with the ``negative emotion'' label.}
\label{fig:cond_prob_negemo}
\end{table}

\begin{table}[h]
    \resizebox{\columnwidth}{!}{
        \begin{tabular}{c|c}
        \toprule 
        \textbf{Feature}              & \textbf{Conditional Probability} \\
        \midrule
        rhetorical question=wh rhq    & 1.0                              \\
        wh question=whq               & 0.946                            \\
        cheeks=tensed                 & 0.903                            \\
        POS=Wh-word                   & 0.881                            \\
        Non-dominant POS=Particle     & 0.776                            \\
        eye brows=lowered             & 0.632                            \\
        eye gaze=to addressee         & 0.583                            \\
        role shift=3                  & 0.570                            \\
        head mvmt: side to side=rapid & 0.562                            \\
        head mvmt: nod=slow           & 0.517                           \\
        \bottomrule
    \end{tabular}
}
\caption{Top ten associated features with the ``confusion'' label.}
\label{fig:cond_prob_confusion}
\end{table}

\subsection{Multi-label Classification Results}
To analyze if we can predict emotionality of the signs using only facial features (FAUs), we train a random forest classifier that takes input of one-hot encodings of facial action units (99 features) for each frame
and predicts one-hot encodings of lexical features (37 features) that are emotion-focused for the sign of each frame. We perform 10-fold cross-validation and the results are shown in Table~\ref{fig:class_results}. We first find heavily imbalanced class labels in the data as indicated by the range of support numbers. To account for the imbalance labels, we calculate micro and weighted f1 average scores to show general performance across all features. We find high precision (0.79) but low recall (0.55), demonstrating that the facial features might be good at capturing emotion-facial relations where there is an obvious correlation, but fails at corner cases where humans express emotions more subtlely.

Looking at each emotion separately, we find that certain emotions such as ``\emph{sad}'' and ``\emph{M-surprise}'' are the easiest to predict while ``\emph{contentment}'' is hard to predict. These seem intuitive since human faces tend to give clear signals when expressing sad or surprise emotions, but contentment is much harder to characterize. However, we also find that ``\emph{horror}'' has 0.0 precision and recall, which might indicate that signers express this emotion with much variety and poses challenges for the classifier to grasp a pattern.

Furthermore, we also rank top 10 FAUs by their importance in the trained random forest classifier shown in Table~\ref{fig:importance_results}. We find that the models mostly deem head positions and movements to have high importance scores. We also find that certain important features match with our previous observation in the co-occurrence analysis. For example, ``\emph{nose wrinkle/tensed}'' and ``\emph{eye brows lowered}'' are also in the top 10 associated features with the ``negative emotion'' and ``confusion'' signs as shown in Tables~\ref{fig:cond_prob_negemo} and~\ref{fig:cond_prob_confusion} and Figures~\ref{fig:negemo_example} and~\ref{fig:confusion_example}. This shows that our findings are consistent across different methods and that FAUs are associated with some emotions strongly.

\begin{table}[h]
\begin{tabular}{lcccc}
\toprule
            feature & precision & recall & f1-score & support \\
\midrule
                 affect &      0.82 &   0.66 &     0.73 &    1539 \\
                  anger &      0.80 &   0.47 &     0.59 &     101 \\
                anxiety &      1.00 &   0.51 &     0.68 &      39 \\
negative emotion (LIWC) &      0.78 &   0.53 &     0.63 &     471 \\
positive emotion (LIWC) &      0.75 &   0.59 &     0.66 &     756 \\
                    sad &      0.88 &   0.97 &     0.92 &      89 \\
              tentative &      0.86 &   0.40 &     0.55 &     189 \\
              affection &      1.00 &   0.27 &     0.42 &      15 \\
             aggression &      0.51 &   0.93 &     0.66 &      40 \\
              confusion &      0.84 &   0.49 &     0.62 &     186 \\
            contentment &      0.57 &   0.08 &     0.14 &      49 \\
                 horror &      0.00 &   0.00 &     0.00 &      73 \\
                   love &      0.80 &   0.15 &     0.26 &      79 \\
negative emotion (Empath) &    0.93 &   0.52 &     0.66 &     234 \\
            nervousness &      0.75 &   0.13 &     0.22 &      70 \\
positive emotion (Empath) &    0.68 &   0.42 &     0.52 &     332 \\
                  shame &      0.75 &   0.13 &     0.22 &      70 \\
               surprise &      0.55 &   1.00 &     0.71 &      29 \\ \hline
              micro avg &      0.79 &   0.55 &     0.65 &    4366 \\
           weighted avg &      0.78 &   0.55 &     0.63 &    4366 \\
\bottomrule
\end{tabular}
\caption{Multi-label classification results on emotional lexicons. We show precision, recall, F1-score, support, and micro/weighted-f1 scores. Generally we find using FAU features to predict emotion labels of a frame gets high precision but low recall. Some emotion features such as ``\emph{sad}'' and ``\emph{surprise}'' are the easiest to predict while ``\emph{contentment}'' is hard to predict.}
\label{fig:class_results}
\end{table}

\begin{table}[h]
\begin{tabular}{lr}
\toprule
                    AU &  Importance \\
\midrule
        head pos: back &    0.101988 \\
  head pos: tilt front &    0.089116 \\
  head pos: turn right &    0.067589 \\
   head pos: tilt left &    0.064259 \\
  head pos: tilt right &    0.063508 \\
   nose wrinkle/tensed &    0.063151 \\
        head mvmt: nod &    0.062214 \\
   head pos: turn left &    0.060699 \\
 eye gaze to addressee &    0.053793 \\
     eye brows lowered &    0.039102 \\
\bottomrule
\end{tabular}
\caption{Top 10 most important facial action units for the random forest classifier on predicting emotionality.}
\label{fig:importance_results}
\end{table}

\section{Discussion}
\paragraph{Co-occurrence Analysis} We find that phrases that have negative valence or traces of confusion do follow a particular pattern, as opposed to something more random (i.e. all features are evenly distributed). Our next steps will be to apply this data to a more complex model, such as finding patterns in sequences of features and predicting sequences of lexical features from sequences of facial features. Specifically, we want to see how emotion features in ASL videos can help with machine translation for ASL by providing affect signals. 

\paragraph{Prediction Task} We find that there is a surprisingly strong relationship between lexical affect and joint facial features. Without any knowledge of what words are being signed, only the face, we predict the emotionality of the signs far more accurately than random guessing. This implies that across signers, there exist stable patterns in the way ASL users use their face to convey meaning. Additionally, it implies that there is a strong relationship between the face and the sign currently being produced, as opposed to matching the face to the emotionality of the entire sentence or showing emotion after the sign is produced. This provides evidence that any ASL recognition system should use facial features as inputs, as they provide a useful signal for meaning.

Our results indicate that \textit{contentment} (e.g. ``pleasure", ``satisfaction", ``happiness") and \textit{horror} (e.g. ``fear", ``terror", ``shock") words were the hardest to predict. Although it is unclear exactly why this occurred, we believe that it is because they are subsets of other categories (e.g. \textit{love} and \textit{anxiety}). Alternatively, it might be the case that these emotions are distinct, but there is no clear convention on how to express them. Both of these hypotheses are supported by the high precision, low recall performance for contentment which indicates that many frames are missed, either due to a competing label or no recognition, but the ones that are found are usually correct.

Among the 10 most important AUs, seven describe head movements (AU IDs 51-60, 83). These, in addition to nose and eyebrow movements, appear to be more meaningful than the lips and mouth, which are usually used to mouth syllables such as in the signs for \textsc{who} (``oo'' mouth), \textsc{large} (``ch'' mouth), and \textsc{awkward} (``th'' mouth). To this extent, they are likely more connected to non-emotional words and therefore not pertinent for the current experiments.

\section{Statement of Contribution}
Lee preprocessed the NCSLGR SignStream data by extracting emotion-focused frames, aligned with facial action units, and conducted statistical analysis on the frames in our dataset. Pei applied the FER model on our frames and conducted analysis on conditional probabilities of different features.

\bibliographystyle{ACM-Reference-Format}
\bibliography{references}

\end{document}